\documentclass[letterpaper, 10 pt, conference]{ieeeconf}  

\IEEEoverridecommandlockouts                              

\usepackage{cite}
\usepackage{multirow}
\usepackage[table,xcdraw]{xcolor}
\usepackage{graphicx}
\usepackage{colortbl} 
\usepackage{amsmath}
\usepackage{dsfont}

\makeatletter
\long\def\@makecaption#1#2{%
  \vskip\abovecaptionskip
  \sbox\@tempboxa{{#1: }#2}%
  \ifdim \wd\@tempboxa >\hsize
    {#1: }#2\par
  \else
    \global \@minipagefalse
    \hb@xt@\hsize{\hfil\box\@tempboxa\hfil}%
  \fi
  \vskip\belowcaptionskip}
\makeatother

\makeatletter
\let\NAT@parse\undefined
\makeatother

\usepackage{caption}
\usepackage{hyperref}

\title{\LARGE \bf
VaLID: Verification as Late Integration of Detections for LiDAR-Camera Fusion
}

\author{Vanshika Vats$^{*}$, Marzia Binta Nizam$^{*}$ and James Davis
\thanks{$^{*}$ Authors contributed equally}
\thanks{Vanshika Vats, Marzia Binta Nizam, and James Davis are with the Department of Computer Science and Engineering, University of California Santa Cruz,
       Santa Cruz, CA 95064, USA
       {\tt\small (vvats, manizam, davisje)@ucsc.edu}}%
}

\begin{document}

\maketitle
\thispagestyle{empty}
\pagestyle{empty}

\begin{abstract}
 Vehicle object detection benefits from both LiDAR and camera data, with LiDAR offering superior performance in many scenarios. Fusion of these modalities further enhances accuracy, but existing methods often introduce complexity or dataset-specific dependencies. In our study, we propose a model-adaptive late-fusion method, VaLID, which validates whether each predicted bounding box is acceptable or not. Our method verifies the higher-performing, yet overly optimistic LiDAR model detections using camera detections that are obtained from either specially trained, general, or open-vocabulary models. VaLID uses a lightweight neural verification network trained with a high recall bias to reduce the false predictions made by the LiDAR detector, while still preserving the true ones. Evaluating with multiple combinations of LiDAR and camera detectors on the KITTI dataset, we reduce false positives by an average of 63.9\%, thus outperforming the individual detectors on 3D average precision (3DAP). Our approach is model-adaptive and demonstrates state-of-the-art competitive performance even when using generic camera detectors that were not trained specifically for this dataset.
\end{abstract}

\section{INTRODUCTION}
The development of autonomous vehicles relies heavily on accurate object detection to safely navigate their environment. These systems use multiple sensors, such as cameras and LiDAR, each offering unique strengths and weaknesses. Cameras provide detailed visual information, while LiDAR offers precise depth information crucial for 3D localization. However, single-sensor approaches have limitations—cameras struggle in low light {\cite{mousavian2017,chabot2017deep}}, and LiDAR lacks high-resolution visual details {\cite{zhou2018,qi2017,shi2019}}.

Multi-modal fusion enhances detection reliability by combining LiDAR depth with camera visuals {\cite{wang2022performance,wang2023multi}}. However, effective integration requires advanced fusion techniques tailored to each modality’s characteristics {\cite{wang2019multi,huang2022multi,zhao2020fusion,arnold2019survey}}. The key challenge is balancing these strengths for robust detection in dynamic environments.

Sensor fusion in autonomous vehicles is categorized into early, deep, and late fusion. Early fusion integrates raw sensor data for detailed interaction but incurs high computational costs due to the need for complex preprocessing steps, like semantic segmentation or depth completion {\cite{vora2020,yin2021,xu2021,wu2023}}. Deep fusion aligns features at a higher level, balancing sensor information but increasing model complexity {\cite{wang2021,mahmoud2023}}. Late fusion, in contrast, merges detection outputs after independent processing, offering flexibility and efficiency but with limited feature interaction {\cite{pang2020,pang2022}}.

While early and deep fusion enhance accuracy by leveraging cross-modal information, late fusion is easier to integrate into industrial workflows, as it allows pre-trained single-modality detectors to be used interchangeably, requiring only detection-level outputs at the time of fusion. An ideal fusion method would combine late fusion’s flexibility with the performance of early and deep fusion.

We introduce a new late fusion method called VaLID: Verification as Late Integration of Detections. This method takes raw detections from a LiDAR model and cross-validates them against corresponding camera detections using a lightweight verification network. Ensuring that only verified LiDAR boxes are kept significantly reduces false positives, improving overall performance.\par

\begin{figure*}[ht!]
  \centering
  \includegraphics[clip, trim=0.15cm 3cm 0cm 0cm, width=\linewidth]{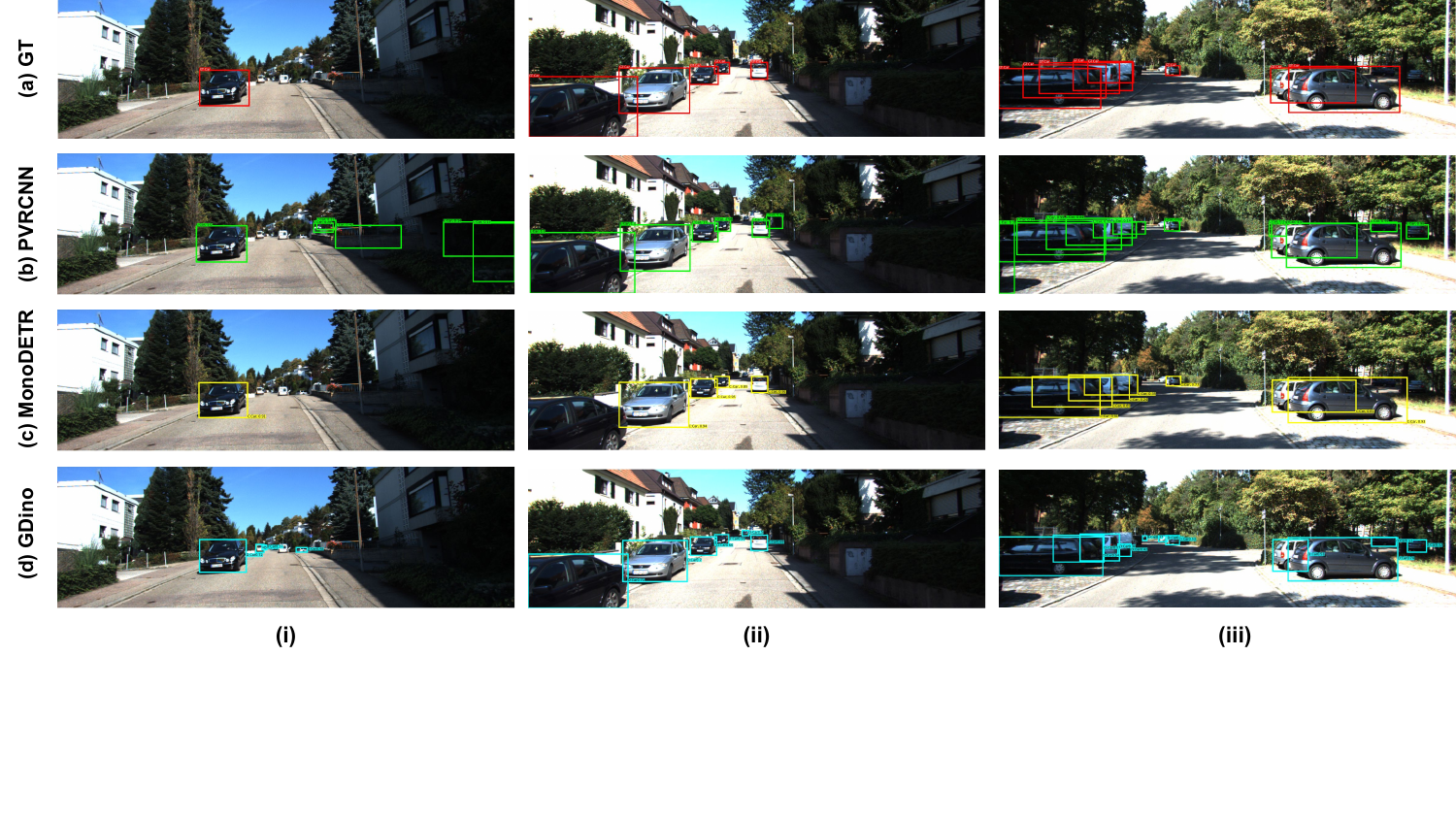}
  \caption{We conduct our study on the KITTI dataset. The figure rows show (a) official 2D ground truth  (b) detections from the specialized LiDAR model PVRCNN, (c) specialized camera model MonoDETR, and (d) open vocabulary model GroundingDino. LiDAR generally produces too many false positives, seen most obviously in Column [i]. Camera models can help verify and reject false positives despite imperfections in their own detections (e.g. Missed detection of partial car in Row[c]Col[ii], and Imperfect dimensions in Row[d]Col[iii]). }
  \label{fig:detections}
\end{figure*}
 
We evaluate our method on the KITTI dataset with two LiDAR detectors, PV-RCNN\cite{shi2020} and TED-S\cite{wu2023TED}. To ensure generality, we use three distinct camera models: MonoDETR\cite{zhang2023monodetr} is a special purpose detector trained specifically on KITTI, YOLO-NAS is a general-purpose model from the YOLO\cite{YOLO} family, and Grounding DINO\cite{liu2023grounding} is an open-vocabulary model.

Dataset-specific models typically outperform general-purpose detectors. Our two LiDAR methods were selected as state-of-the-art methods designed and tested on the KITTI dataset, while general-purpose camera models were included for broader evaluation. Though these models perform worse as primary detectors, our method leverages them for the simpler task of verification.

We find that the VaLID method removes an average 63.9\% of false positives, consistently improving overall detection average precision. 
Importantly, the improvement is as effective using the general camera models as it is with the specifically trained method. In both cases, results are competitive with the state of the art.

Thus, the primary contributions of our work are:
\begin{itemize}
    \item We propose VaLID, a lightweight, model-adaptive late-fusion approach that focuses on reducing false positives by cross-validating them with camera detections.
    \item Our method works with both specialized and general-purpose camera models without requiring individual model retraining or dataset-specific fine-tuning, making it adaptable to various detection backbones.
    \item Our method uses a learnable verification module with confidence rescoring that balances simplicity and effectiveness, avoiding the complexity of deep-fusion methods while achieving competitive performance.
\end{itemize}

\section{Related Work}
\subsection{Detection Using Camera Images}

Cameras images are fundamentally a 2D data source, and 2D object detection is well-studied with general-purpose and domain-specific datasets~\cite{wang2023comprehensive,liu2023grounding,liu2016ssd,shehzadi20232d,lin2014microsoft}.  Specialized models trained on vehicle datasets achieve higher accuracy {\cite{geiger2013vision,sun2020scalability,caesar2020nuscenes}}, but autonomous driving requires 3D awareness. Depth estimation from images alone is challenging—methods like Deep3DBox {\cite{mousavian2017}}, MonoDETR {\cite{zhang2023monodetr}}, and MonoPGC {\cite{wu2023monopgc}} refine depth, while Pseudo-LiDAR {\cite{wang2019pseudo}} simulates LiDAR but increases computational cost. Multi-task models like Deep MANTA {\cite{chabot2017deep}} improve efficiency but struggle with depth reliability. Camera-only methods remain limited in 3D accuracy without additional sensors. Unfortunately, camera-only methods, while effective in some scenarios, are limited by their inability to produce reliable depth estimation without additional sensors, and 3D bounding box accuracy remains below 2D bounding box accuracy. In our work, we use the higher-accuracy 2D detections from camera images as one of the inputs to our late fusion method along with the LiDAR information.

\subsection{Detection Using LiDAR Point Clouds}
LiDAR-based 3D detection has become a fundamental component in autonomous driving due to its ability to capture detailed spatial information. Early one-stage models like PointNet and PointNet++ processed raw point clouds {\cite{qi2017,qi2017pointnet++}}, while range-image methods {\cite{meyer2019,bewley2020}} encoded depth at the pixel-level but faced real-time and distance limitations.

Two-stage models improved accuracy by introducing 3D voxels and sparse convolutions, as seen in VoxelNet {\cite{zhou2018}} and SECOND {\cite{yan2018}}. PV-RCNN {\cite{shi2020}} combines voxel-based CNNs with PointNet for high-quality proposals, while PointRCNN {\cite{shi2019}} and Part-A2Net {\cite{shi2020}} further refine detections. Although LiDAR provides robust spatial data, it struggles with complex and distant objects. In our late fusion approach, LiDAR detections are combined with camera detections to provide more accurate and reliable 3D object detection.

\subsection{Detection Using LiDAR-Camera Fusion}
Multi-modal fusion techniques for 3D detection are increasingly common, particularly in autonomous driving, where camera and LiDAR sensors complement each other. These techniques are categorized into three main approaches: early, deep, and late fusion.\par 

Early fusion integrates camera data with LiDAR early in the pipeline to enhance detection. Methods like PointPainting and MVP enrich LiDAR by projecting 2D segmentation or generating virtual points to mitigate sparsity {\cite{vora2020,yin2021}}, while FusionPainting refines this using 3D segmentation {\cite{xu2021}}. Techniques like VirConv apply virtual points and Stochastic Voxel Discard to reduce noise but add computational overhead {\cite{wu2023,dou2019}}.\par

\begin{figure}[ht!]
  \centering
  \includegraphics[clip, trim=0cm 0cm 0cm 0cm, width=0.88\linewidth]{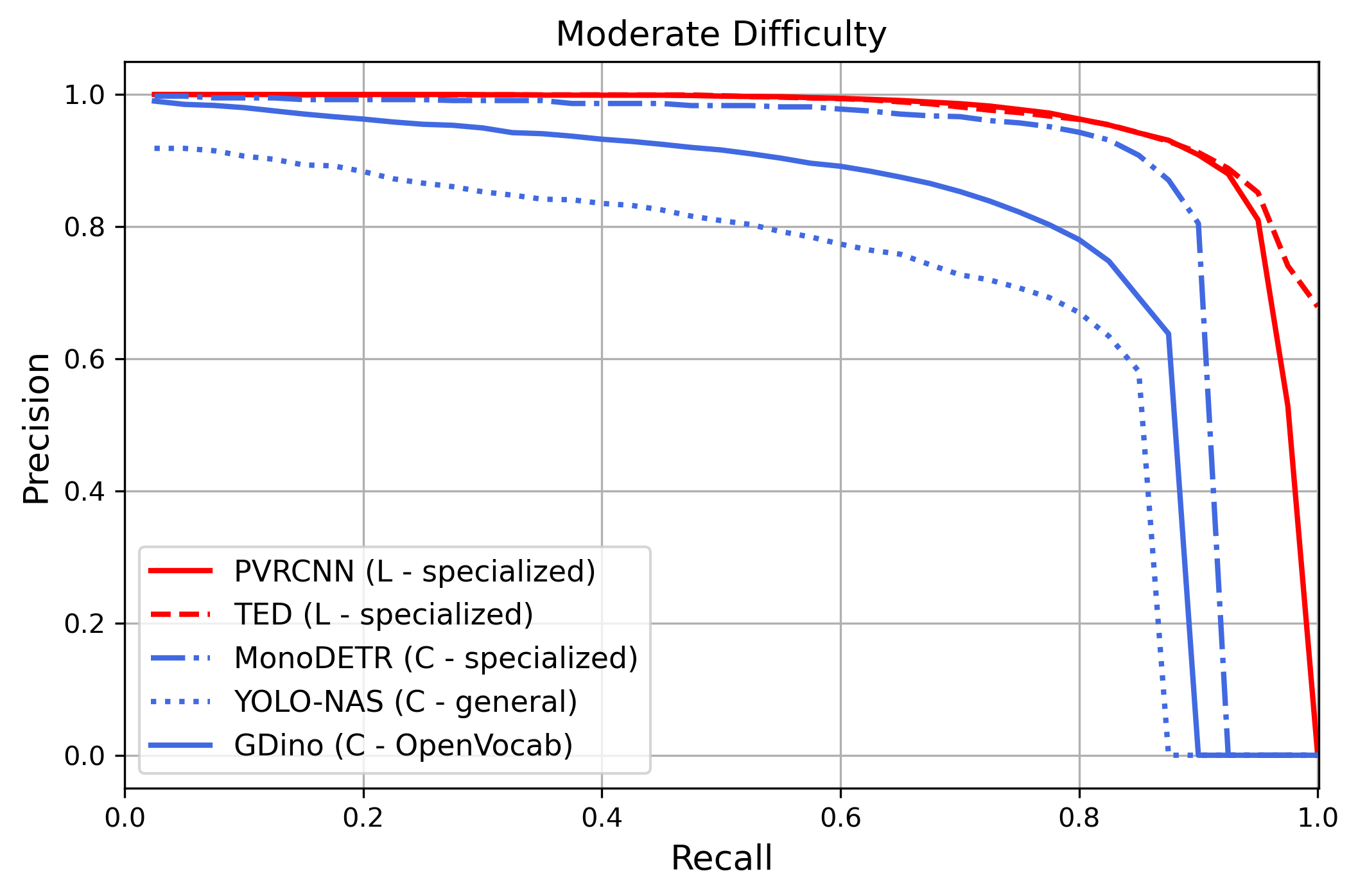}
  \vspace{-0.3cm}
  \caption{Precision-Recall curve (PRC) of our baseline single modality models on the KITTI moderate difficulty data set. Notice that both LiDAR models outperform all three of the camera models and that the specialized camera detector outperforms the two general-purpose camera-based object detectors.}
  \label{fig:prc_all}
\end{figure}

Deep fusion aligns features from both modalities at a higher level. PointAugmenting and DVF bridge the 2D-3D gap by projecting point clouds onto image features or reweighting voxels {\cite{wang2021,mahmoud2023}}. Attention-based models like EPNet, EPNet++, and DeepFusion dynamically refine feature fusion {\cite{huang2020,liu2022,li2022}}. SFD fuses features in the 3D RoI, while BEVFusion unifies them into a Bird’s Eye View (BEV) {\cite{wu2022,liu2023}}. \par

Late fusion combines sensor outputs after independent processing. Methods like CLOCs and Fast-CLOCs refine detection confidence by ensuring 2D-3D box alignment but rely on raw sensor data, limiting feature alignment {\cite{pang2020,pang2022}}. C-CLOCs work improves this by using contrastive learning to align 2D-3D features, enhancing accuracy in challenging environments {\cite{zhang2024}}. Another approach integrates 2D segmentation with 3D LiDAR to reduce false positives from roadblocks and tunnel walls {\cite{ccaldiran2022}}. All of these late-fusion methods for vehicle detection have focused on combining specialized LiDAR and camera vehicle detectors.

Our work proposes a new late fusion method that is lightweight, model-adaptive, and competitive with the state of the art. We demonstrate results with both a specialized camera detector trained on the KITTI dataset and with more general vision models that lack specific knowledge of the dataset.


\section{Background}
\subsection{LiDAR and Camera Models for Late-Fusion}
Many urban driving object detection methods utilize LiDAR and camera modalities separately, as well as in fusion, for enhanced performance. Most  methods, however, use specialized training and fine-tuning tailored to specific datasets, which is both time-consuming and computationally demanding. To reduce this dependency, our study leverages general and open-vocabulary camera models without requiring dataset-specific fine-tuning. We perform our study on the KITTI dataset~\cite{geiger2013vision}, using PVRCNN \cite{shi2020} and TED-S \cite{wu2023TED} models as the primary LiDAR models. For fusion, we use MonoDETR \cite{zhang2023monodetr} as a specialized KITTI model, and YOLO-NAS \cite{YOLO} as a general camera detector that is not trained on KITTI. GroundingDINO \cite{liu2023grounding} is used as an open-vocabulary model that can detect objects that are not even seen during the training. Although not as precise as fine-tuned models, these general models offer cross-domain versatility, providing the flexibility needed for a more robust late-fusion approach. 

\begin{figure}[ht!]
  \centering
  \includegraphics[clip, trim=0cm 2.7cm 16cm 0cm, width=0.88\linewidth]{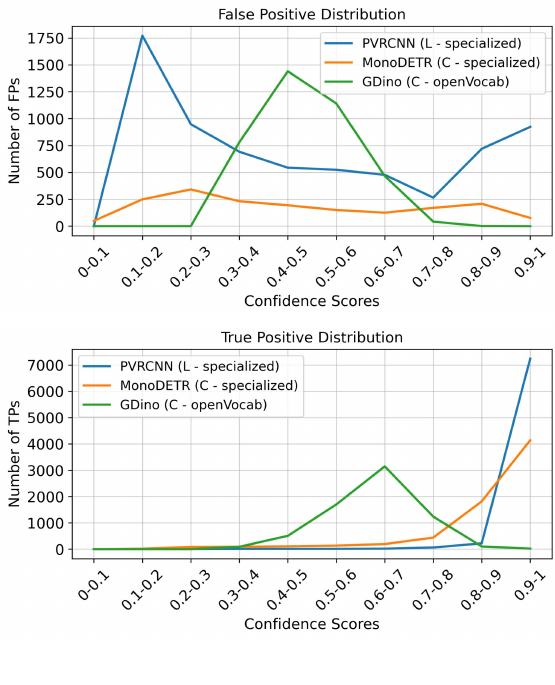}
  \vspace{-0.7cm}
  \caption{The distribution of false positives and true positives across confidence score bands for one LiDAR and two camera object detection models on the KITTI moderate set. Notice that the LiDAR method has false positives across all confidence bands, and that the camera methods have different and sometimes complementary distributions.}
  \label{fig:tp_fp_dist}
\end{figure}

\subsection{Model Detections Complement Each Other}
LiDAR and camera models have their own strengths and weaknesses.  Fig. \ref{fig:detections} illustrates how these models, while missing certain detections, often compensate for one another's limitations. For instance, PV-RCNN detects many objects but generates false positives, while MonoDETR and GroundingDINO miss some detections yet identify objects PV-RCNN overlooks. GroundingDINO, despite its imperfect boxes and false positives, detects objects missed by specialized models.

Fig.~\ref{fig:prc_all} shows the overall performance of our baseline LiDAR and camera methods. Notice that both the LiDAR methods outperform all three camera methods, and that the specialized camera model outperforms the two general models. Since the performance of LiDAR methods exceeds the camera methods, we take them as our primary detectors and use the camera methods to augment their detection ability.

Unfortunately, the tested LiDAR methods produce many false positives which negatively impacts the overall system performance. Reducing these false positives is crucial for enhancing precision. To better understand the distribution of false positives, Fig.~\ref{fig:tp_fp_dist} shows false positives and true positives over different confidence bands for several methods. Notice that the overall distributions are different in different methods. The neural network in our fusion model learns to leverage these differences.

\begin{figure*}[ht!]
  \centering
  \includegraphics[clip, trim=0cm 7cm 7.5cm 0cm, width=0.82\linewidth]{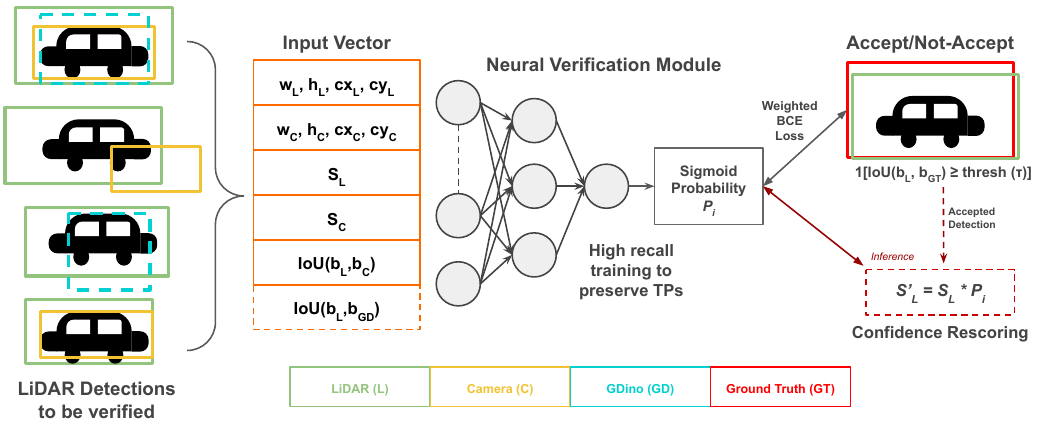}
  \vspace{-0.2cm}
  \caption{In our proposed method, VaLID, we take LiDAR ($b_L$) boxes as the primary detections and use the camera ($b_C$) modality to verify whether each $b_L$ is an acceptable detection or not. The input vector is represented as the bounding box dimensions of width $w$, height $h$, and center $(cx, cy)$, confidence scores $s$, and the measure of overlap $IoU$.  We pass this vector through a neural verification module which outputs an acceptance probability as a sigmoid value. The training objective is defined based on the overlap between the ground truth ($b_{GT}$) with the given $b_L$, optimized using a weighted BCE loss to encourage high recall. During inference, the accepted box confidence is rescored by multiplying it with the output sigmoid probability in order to reduce the confidence of the lingering false positives.}
  \label{fig:methodology}
\end{figure*}

\section{Design and Implementation} 
\subsection{Method Overview}
\label{sec:methods}
We use camera model detections and a neural module to verify whether a particular LiDAR bounding box detection is valid. Fig.~\ref{fig:methodology} provides an overview of our method.
For each LiDAR detection, we look for a corresponding camera detection above a certain overlap score (see Sec. \ref{subsec:implementation}). If a matching camera box is found, the camera bounding box, the Intersection over Union (IoU) between the LiDAR and camera boxes, and the confidence scores of both detections are fed into a neural network, along with the LiDAR box. If no matching camera box is found, we input a zero for both the bounding box and the confidence score for the camera model. 
The inputs given would be:
\begin{equation} \label{eq:1}
 I = \{ b_L, b_C, S_L, S_C, IoU(b_L, b_{C}), [IoU(b_L, b_{GD})] \}
\end{equation}
\noindent where ${b}_L$ and ${b}_C$ are the dimensions of bounding boxes of LiDAR and camera models respectively, represented by their width, height, and center coordinates, normalized by the image's dimensions. ${S}_L$ and ${S}_C$ are their confidence scores, and $IoU(b_L, b_{C})$ is the Intersection over Union of LiDAR and camera detection. When two camera methods are used, information about a second bounding box is provided. For example, $IoU(b_L, b_{GD})$ is the optionally provided IoU between LiDAR and GroundingDINO.

Our goal is to determine whether to keep or discard each LiDAR (${b}_L$) detection. We generate ground truth by calculating the IoU between the LiDAR box and the official KITTI ground truth ${b}_{\mathcal{GT}}$  and an IoU threshold $\tau$ as per the dataset's original settings. Detections above this threshold are marked as valid (positive class), while those below are ignored. 
\vspace{-0.2cm}
\begin{equation} \label{eq:2}
   GT_{VaLID} = \mathds{1}[\operatorname{IoU}(b_L, b_{\mathcal{GT}}) \geq \tau]
\end{equation}
where $\mathds{1}[\cdot]$ is the indicator function, which returns 1 if the $IoU(b_L, b_{\mathcal{GT}})$ exceeds a threshold of $\tau$ = 0.7 for the \textit{Car} class; and 0 otherwise, i.e., rejected.

Validating LiDAR detections with camera detections could be approached with a simple thresholding heuristics. However, simple heuristics would fail to capture the complex, non-linear relationships between LiDAR and camera detections, such as bounding boxes with high confidence but low IoU or cases where one modality compensates for the other.
A neural network allows this relationship to be learned from the data rather than hand-tweaked by the researcher for each new pair of detectors. Given the small, fixed-size feature vector in a structured format, we employ a lightweight neural verification module that learns to validate LiDAR detections by leveraging spatial and confidence-based cues from complementary camera observations. This verification network is compact and ingests a structured feature vector comprising geometric attributes, confidence scores, and inter-modal spatial alignment metrics (IoU). It can learn non-linear patterns and adapt to biases in false positives across different parametric bands, which are not the same in different detection models. More complex models like CNNs or transformers, designed for high-dimensional inputs, would add unnecessary complexity without offering meaningful improvements. 

The input vector \textit{I} (Eq.~\ref{eq:1}) is, therefore, fed to a three-layered fully-connected feedforward verification module that is specifically designed to address the tradeoff between preserving true positives and reducing false positives.

\begin{equation} \label{eq:3}
    P_i = \sigma(f_{\theta}(I_i))
\end{equation}

where $P_i$ is the sigmoid $\sigma$ probability score for the learned function $f_{\theta}$. The module is trained with a high recall bias to minimize the risk of mistakenly discarding true positive detections while effectively filtering out false positives. Formally, we aim to minimize the false positives $(b_i \notin \mathcal{GT})$ while maintaining high recall over the number of LiDAR detections $N_L$ by optimizing, 
\begin{equation} \label{eq:4}
\min_{\theta} \sum_{i=1}^{N_L} \left( P_i \mathds{1}[b_i \notin \mathcal{GT}] + \lambda (1 - P_i) \mathds{1}[b_i \in \mathcal{GT}] \right)
\end{equation}
where $\mathds{1}[b_i \in GT]$ represents true positives. Thus, the first half of Eq. \ref{eq:4} minimizes false positives by discouraging high-confidence incorrect detections, and the second half penalizes false negatives by a factor of $\lambda$ (Sec. \ref{subsec:implementation}), ensuring true positives retain high confidence.



This module is architecturally identical in all of our experiments. However, the activation weights need to be trained for each pair of LiDAR and camera methods since each baseline method has its own distributions of true and false positives, and the goal of this verification is to make specific tradeoffs in these distributions.

To ensure that enough verification boxes exist, we desire an over-production of bounding boxes in the camera-based verification methods as opposed to under-production, and thus include even boxes with low confidence. We observe that the open vocabulary camera model provides detected boxes that are not as precisely aligned as the specialized LiDAR and camera models, especially around occluded objects, and thus adjust the IoU threshold to ensure sufficient matching boxes exist. 

Since we employ high-recall training, we expect some false positives will continue to be accepted by the verification module during inference. To mitigate their impact on the overall average precision (AP), we assign a final confidence ($S_{L}'$) to each box by multiplying the accepted LiDAR box confidence score $S_{L}$ with the predicted output sigmoid probability $P_{i}$. 
\begin{equation}
    S'_{L} = S_L * P_i
\end{equation}
This ensures that true positive boxes retain high confidence, while false positives with lower predicted probabilities are down-weighted, effectively balancing the AP.



\begin{table}[]
\renewcommand{\arraystretch}{1.2}
\centering
\resizebox{\columnwidth}{!}{%
\begin{tabular}{|l|c|rr|rr|}
\hline
\multicolumn{1}{|c|}{} &     & \multicolumn{2}{c|}{\textbf{Easy}}                       & \multicolumn{2}{c|}{\textbf{Hard}}                       \\ \cline{3-6} 
\multicolumn{1}{|c|}{\multirow{-2}{*}{\textbf{Models}}} &
  \multirow{-2}{*}{\textbf{Modality}} &
  \multicolumn{1}{c|}{\textbf{TP}} &
  \multicolumn{1}{c|}{\textbf{FP}} &
  \multicolumn{1}{c|}{\textbf{TP}} &
  \multicolumn{1}{c|}{\textbf{FP}} \\ \hline
\rowcolor[HTML]{F3F3F3} 
PVRCNN                 & L   & \multicolumn{1}{r|}{\cellcolor[HTML]{F3F3F3}1433} & 1287 & \multicolumn{1}{r|}{\cellcolor[HTML]{F3F3F3}5364} & 3620 \\
PV+Mono (Ours)               & LC  & \multicolumn{1}{r|}{1428}                         & 393  & \multicolumn{1}{r|}{5280}                         & 1084 \\
PV+YOLO (Ours)                & LC  & \multicolumn{1}{r|}{1430}                         & 360  & \multicolumn{1}{r|}{5278}                         & 965  \\
PV+GDino (Ours)               & LC  & \multicolumn{1}{r|}{1430}                         & 366  & \multicolumn{1}{r|}{5271}                         & 937  \\
PV+Mono+GDino (Ours)          & LCC & \multicolumn{1}{r|}{1430}                         & 354  & \multicolumn{1}{r|}{5261}                         & 906  \\
PV+YOLO+GDino (Ours)          & LCC & \multicolumn{1}{r|}{1430}                         & 347  & \multicolumn{1}{r|}{5266}                         & 887  \\ \hline
\rowcolor[HTML]{F3F3F3} 
TED                    & L   & \multicolumn{1}{r|}{\cellcolor[HTML]{F3F3F3}1434} & 864  & \multicolumn{1}{r|}{\cellcolor[HTML]{F3F3F3}5425} & 1848 \\
TED+Mono (Ours)               & LC  & \multicolumn{1}{r|}{1434}                         & 323  & \multicolumn{1}{r|}{5210}                         & 599  \\
TED+YOLO (Ours)               & LC  & \multicolumn{1}{r|}{1434}                         & 389  & \multicolumn{1}{r|}{5380}                         & 1028 \\
TED+GDino (Ours)              & LC  & \multicolumn{1}{r|}{1434}                         & 352  & \multicolumn{1}{r|}{5359}                         & 869  \\
TED+Mono+GDino (Ours)         & LCC & \multicolumn{1}{r|}{1434}                         & 334  & \multicolumn{1}{r|}{5343}                         & 826  \\
TED+YOLO+GDino (Ours)         & LCC & \multicolumn{1}{r|}{1434}                         & 381  & \multicolumn{1}{r|}{5354}                         & 914  \\ \hline
\end{tabular}%
}
\caption{Since ours is a late fusion method that generates no new detections, we focus on reducing false positives (FPs) while preserving  true positives (TPs). Our method is successfully able to do so, reducing the FPs by an average of 73.6\% on PVRCNN and 54.2\% on TED.}
\label{tab:fp}
\end{table}

\begin{figure}[ht!]
  \centering
  \includegraphics[clip, trim=0cm 0cm 14cm 0cm, width=0.85\linewidth]{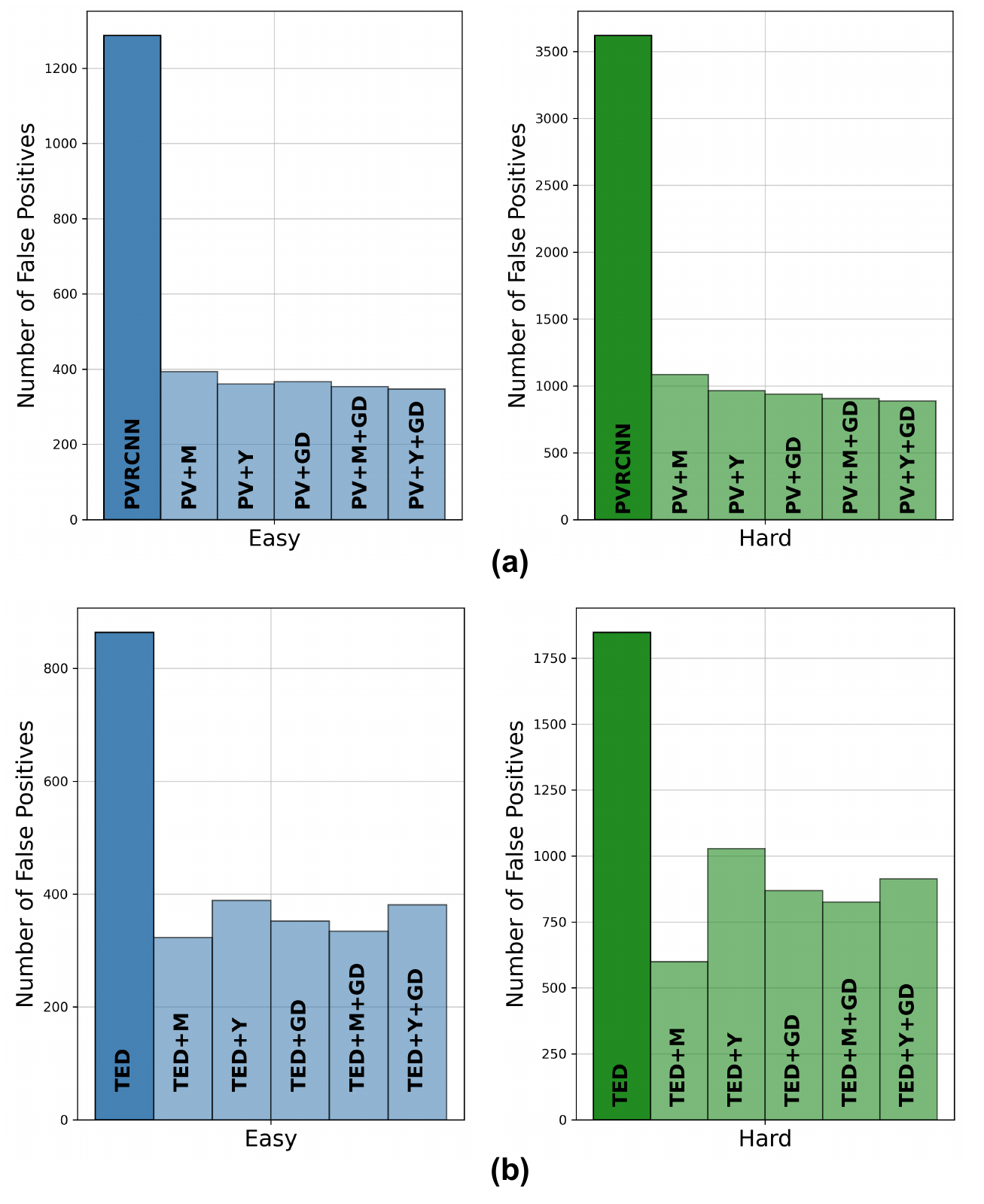}
  \vspace{-0.3cm}
  \caption{Our fusion method is able to reduce the false positives significantly on (a) PVRCNN, and (b) TED model detections. Note that this improvement is achieved for all tested camera methods used for verification. }
  \label{fig:fp}
\end{figure}

\subsection{Implementation}
\label{subsec:implementation}
To implement our late-fusion method, we test two specialized LiDAR detectors (PVRCNN and TED) in combination with multiple camera models: a specialized model (MonoDETR), a general-purpose model (YOLO-NAS), and an open-vocabulary model (GroundingDINO). The experiments are conducted on the KITTI dataset. We remove dataset images for which ground truth is not available, as well as those that have already been used to train our baseline models. This leaves 3769 images available for our experiments comprising the subset that the dataset providers call validation. We split these images into training and testing subsets. For PVRCNN, these images generate a training set of 12,303 LiDAR bounding box samples and test set of 12,817 bounding boxes. For TED-S, we obtain 9,016 training and 9,406 testing  boxes.

We match the initial LiDAR and corresponding camera boxes with an IoU of 0.5 or higher. This threshold is intentionally set lower than the official KITTI benchmark of 0.7 to account for potential inaccuracies in the camera detection dimensions, ensuring that valid detections are not inadvertently discarded due to minor misalignments. The final inputs pass through a three-layer fully connected feedforward neural module with ReLU \cite{relu} activations, followed by a sigmoid output for probabilistic confidence refinement. To prioritize reducing false positives while preserving true positives, we apply a weighting factor $\lambda = 10$ to penalize false negatives more strongly than false positives. This corresponds to a class weighting of 1:10 for negative and positive classes, ensuring that the model maintains high recall while filtering incorrect detections. The model is thus optimized using a weighted binary cross-entropy loss function, which compares the sigmoid outputs against the ground truth labels to predict the acceptance or rejection of a given LiDAR detection. We train our models for 50 epochs with an Adam learning rate of 0.0001. During inference, we perform confidence rescoring, refining the final detection confidence.


\begin{table*}[]
\renewcommand{\arraystretch}{1.45}
\scriptsize
\centering
\resizebox{0.9\textwidth}{!}{%
\begin{tabular}{|l|c|c|rrr|}
\hline
\multicolumn{1}{|c|}{} &
   &
   &
  \multicolumn{3}{c|}{\textbf{3D AP / 2D AP (\%)}} \\ \cline{4-6} 
\multicolumn{1}{|c|}{\multirow{-2}{*}{\textbf{Models}}} &
  \multirow{-2}{*}{\textbf{Modality}} &
  \multirow{-2}{*}{\textbf{Model Type}} &
  \multicolumn{1}{c|}{\textbf{Easy}} &
  \multicolumn{1}{c|}{\textbf{Moderate}} &
  \multicolumn{1}{c|}{\textbf{Hard}} \\ \hline
MonoDETR\cite{zhang2023monodetr} &
  C &
  Specialized &
  \multicolumn{1}{r|}{28.05 / 95.12} &
  \multicolumn{1}{r|}{20.77 / 87.36} &
  17.43 / 79.94 \\
YOLO-NAS\cite{YOLO} &
  C &
  General &
  \multicolumn{1}{r|}{N/A / 75.32} &
  \multicolumn{1}{r|}{N/A / 68.69} &
  N/A / 55.11 \\
GroundingDino\cite{liu2023grounding} &
  C &
  Open-Vocab &
  \multicolumn{1}{r|}{N/A / 84.73} &
  \multicolumn{1}{r|}{N/A / 78.68} &
  N/A / 62.82 \\ \hline
\rowcolor[HTML]{CCCCCC} 
\textbf{PVRCNN}\cite{shi2020} &
  \textbf{L} &
  Specialized &
  \multicolumn{1}{r|}{\cellcolor[HTML]{CCCCCC}91.94 / 98.30} &
  \multicolumn{1}{r|}{\cellcolor[HTML]{CCCCCC}82.97 / 94.44} &
  82.44 / 94.08 \\ \hline
\rowcolor[HTML]{EFEFEF} 
Ours \emph{(PVRCNN+MonoDETR)} &
  LC &
  Late-Fusion &
  \multicolumn{1}{r|}{\cellcolor[HTML]{FFF6DE}91.96 / 98.91} &
  \multicolumn{1}{r|}{\cellcolor[HTML]{FFF6DE}83.47 / 95.45} &
  \cellcolor[HTML]{FFF6DE}82.89 / 93.11 \\
\rowcolor[HTML]{EFEFEF} 
Ours \emph{(PVRCNN+YOLO-NAS)} &
  LC &
  Late-Fusion &
  \multicolumn{1}{r|}{\cellcolor[HTML]{FFF2CC}92.30 / 99.09} &
  \multicolumn{1}{r|}{\cellcolor[HTML]{FFE9AA}83.83 / 95.83} &
  \cellcolor[HTML]{FFF2CC}82.91 / 93.33 \\
\rowcolor[HTML]{EFEFEF} 
Ours \emph{(PVRCNN+GDino)} &
  LC &
  Late-Fusion &
  \multicolumn{1}{r|}{\cellcolor[HTML]{FFDB71}92.41 / 98.91} &
  \multicolumn{1}{r|}{\cellcolor[HTML]{FFF2CC}83.54 / 95.26} &
  \cellcolor[HTML]{FFDB71}83.01 / 93.01 \\
\rowcolor[HTML]{EFEFEF} 
Ours \emph{(PVRCNN+MonoDETR+GDino)} &
  LCC &
  Late-Fusion &
  \multicolumn{1}{r|}{\cellcolor[HTML]{FFE9AA}92.37 / 99.15} &
  \multicolumn{1}{r|}{\cellcolor[HTML]{FFDB71}83.91 / 95.87} &
  \cellcolor[HTML]{FFCB32}83.08 / 93.37 \\
\rowcolor[HTML]{EFEFEF} 
Ours \emph{(PVRCNN+YOLO-NAS+GDino)} &
  LCC &
  Late-Fusion &
  \multicolumn{1}{r|}{\cellcolor[HTML]{FFCB32}92.46 / 99.24} &
  \multicolumn{1}{r|}{\cellcolor[HTML]{FFCB32}83.93 / 95.99} &
  \cellcolor[HTML]{FFE9AA}82.95 / 93.42 \\ \hline
\rowcolor[HTML]{CCCCCC} 
\textbf{TED-S}\cite{wu2023TED} &
  \textbf{L} &
  Specialized &
  \multicolumn{1}{r|}{\cellcolor[HTML]{CCCCCC}95.23 / 98.88} &
  \multicolumn{1}{r|}{\cellcolor[HTML]{CCCCCC}87.98 / 96.73} &
  86.28 / 94.95 \\ \hline
\rowcolor[HTML]{EFEFEF} 
Ours \emph{(TED+MonoDETR)} &
  LC &
  Late-Fusion &
  \multicolumn{1}{r|}{\cellcolor[HTML]{FFF2CC}95.40 / 99.23} &
  \multicolumn{1}{r|}{\cellcolor[HTML]{FFF6DE}88.69 / 95.64} &
  \cellcolor[HTML]{FFF2CC}86.77 / 95.59 \\
\rowcolor[HTML]{EFEFEF} 
Ours \emph{(TED+YOLO-NAS)} &
  LC &
  Late-Fusion &
  \multicolumn{1}{r|}{\cellcolor[HTML]{F3F3F3}95.18 / 99.12} &
  \multicolumn{1}{r|}{\cellcolor[HTML]{FFF2CC}88.73 / 95.79} &
  \cellcolor[HTML]{FFF6DE}86.64 / 95.66 \\
\rowcolor[HTML]{EFEFEF} 
Ours \emph{(TED+GDino)} &
  LC &
  Late-Fusion &
  \multicolumn{1}{r|}{\cellcolor[HTML]{FFDB71}95.86 / 99.35} &
  \multicolumn{1}{r|}{\cellcolor[HTML]{FFDB71}88.85 / 95.69} &
  \cellcolor[HTML]{FFE9AA}86.78 / 95.57 \\
\rowcolor[HTML]{EFEFEF} 
Ours \emph{(TED+MonoDETR+GDino)} &
  LCC &
  Late-Fusion &
  \multicolumn{1}{r|}{\cellcolor[HTML]{FFE9AA}95.59 / 99.40} &
  \multicolumn{1}{r|}{\cellcolor[HTML]{FFCB32}88.94 / 95.93} &
  \cellcolor[HTML]{FFDB71}86.80 / 95.74 \\
\rowcolor[HTML]{EFEFEF} 
Ours \emph{(TED+YOLO-NAS+GDino)} &
  LCC &
  Late-Fusion &
  \multicolumn{1}{r|}{\cellcolor[HTML]{FFF6DE}95.38 / 99.19} &
  \multicolumn{1}{r|}{\cellcolor[HTML]{FFE9AA}88.78 / 95.87} &
  \cellcolor[HTML]{FFCB32}86.83 / 95.70 \\ \hline
CLOCs \cite{pang2020} &
  LC &
  Late-Fusion &
  \multicolumn{1}{r|}{92.74 / 99.33} &
  \multicolumn{1}{r|}{82.90 / 93.75} &
  77.75 / 92.89 \\
CMKD\cite{CMKD} &
  C(+L) &
  Knowledge Distil. &
  \multicolumn{1}{r|}{34.56 / 96.08} &
  \multicolumn{1}{r|}{23.10 / 92.43} &
  19.52 / 87.11 \\
  LogoNet\cite{logonet} &
  LC &
  Deep Fusion &
  \multicolumn{1}{r|}{92.04 / 97.04} &
  \multicolumn{1}{r|}{85.04 / 89.44} &
  84.31 / 89.20 \\
\hline
\end{tabular}%
}
\caption{Perfomance on the Average Precision (AP) metric for our method on all difficulty sets on KITTI-val for the \emph{Car} class. In general, the baseline camera methods do not produce 3D bounding boxes (N/A) or produce ones with very low performance (MonoDETR). Our late fusion method consistently outperforms the original specialized LiDAR and camera methods in 3DAP across all the difficulty levels. Our method is also comparable to the SOTA deep fusion methods, despite being simpler and lightweight, without requiring special integrated training on the individual models. The color coding in yellow emphasizes the degree of improvement on 3DAP (primary comparison metric) over the baselines.}
\label{table:main}
\end{table*}

\begin{figure*}[ht!]
  \centering
  \includegraphics[clip, trim=0cm 4.5cm 0cm 0cm, width=\linewidth]{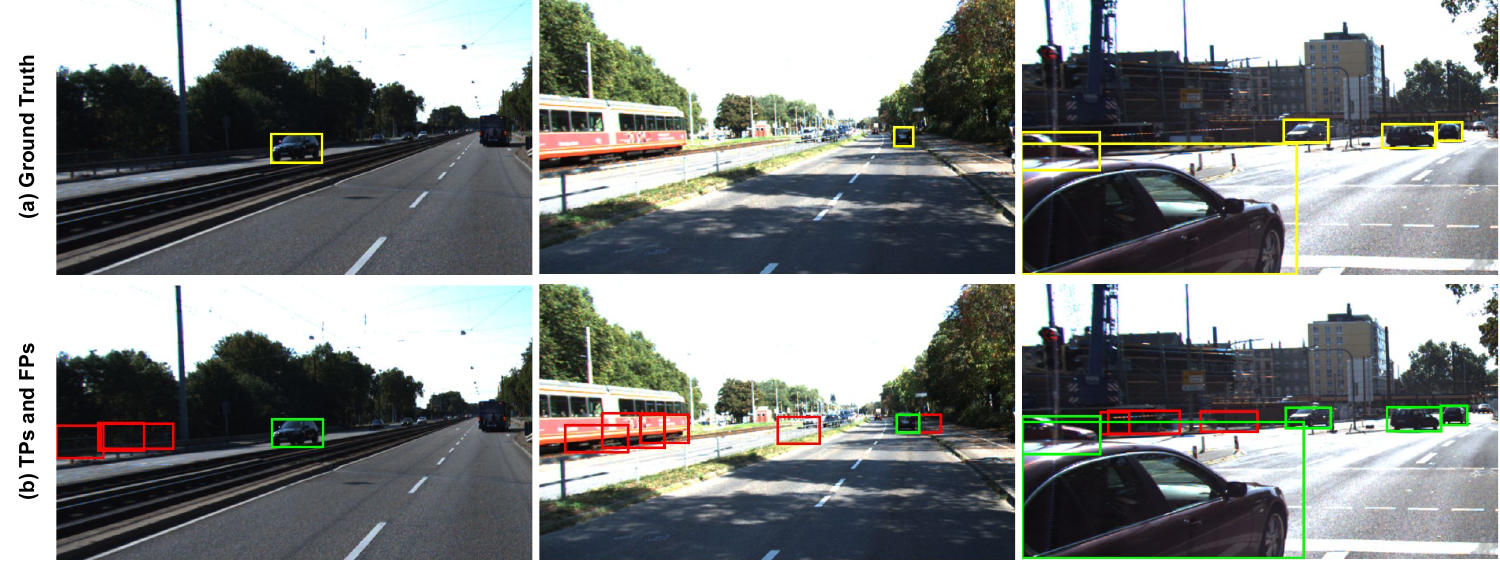}
  \caption{Qualitative results illustrating the reduction of false positives from our fusion method as compared to the specialized LiDAR PVRCNN. (a) denotes the ground truth bounding boxes in the KITTI-val set. In (b), red boxes represent the false positives generated by PVRCNN, while green boxes indicate the final accepted boxes from our late fusion method, demonstrating a significant decrease in false positives.}
  \label{fig:qualitative}
\end{figure*}

\section{Results}

We present the results of our methods in two parts: First, we show the impact of our fusion method on reducing false positives generated by LiDAR-only models. Second, we provide an overall comparison of the Average Precision (3D/2DAP), comparing our fusion method with single-modality models as well as state-of-the-art fusion methods.

\subsection{False Positives Reduction}
Table \ref{tab:fp} shows the performance of LiDAR-only models in terms of true positive and false positive bounding box predictions. Note that both LiDAR models produce a large number of false positives. Our late fusion method is evaluated on each LiDAR model, using each of our three baseline camera detectors, as well as using combinations of camera detectors. Our method consistently and substantially reduces false positives across all combinations. This is made especially clear through the visualization in Figure \ref{fig:fp}.  Our late fusion method achieves an average false positive reduction of 73.6\% on PVRCNN and 54.2\% on TED. Importantly, the true positive (TP) numbers have only a small reduction, indicating that the significant decrease in false positives does not come at the cost of many missed detections. The results are qualitatively shown in Figure \ref{fig:qualitative}.

\begin{table}[]
\renewcommand{\arraystretch}{1.5}
\scriptsize
\centering
\resizebox{\columnwidth}{!}{%
\begin{tabular}{|l|rrr|}
\hline
\multicolumn{1}{|c|}{} &
  \multicolumn{3}{c|}{\textbf{3D AP / 2D AP (\%)}} \\ \cline{2-4} 
\multicolumn{1}{|c|}{\multirow{-2}{*}{\textbf{Models}}} &
  \multicolumn{1}{c|}{\textbf{Easy}} &
  \multicolumn{1}{c|}{\textbf{Moderate}} &
  \multicolumn{1}{c|}{\textbf{Hard}} \\ \hline
\rowcolor[HTML]{CCCCCC} 
\textbf{PVRCNN}\cite{shi2020} &
  \multicolumn{1}{r|}{\cellcolor[HTML]{CCCCCC}64.25 / 72.00} &
  \multicolumn{1}{r|}{\cellcolor[HTML]{CCCCCC}57.00 / 66.58} &
  52.00 / 63.34 \\ \hline
\rowcolor[HTML]{EFEFEF} 
Ours \emph{(PV+YOLO)} &
  \multicolumn{1}{r|}{\cellcolor[HTML]{FFE9AA}67.09 / 73.54} &
  \multicolumn{1}{r|}{\cellcolor[HTML]{FFE9AA}59.79 / 68.57} &
  \cellcolor[HTML]{FFE9AA}53.74 / 64.69 \\
\rowcolor[HTML]{EFEFEF} 
Ours \emph{(PV+GDino)} &
  \multicolumn{1}{r|}{\cellcolor[HTML]{FFF2CC}65.64 / 72.02} &
  \multicolumn{1}{r|}{\cellcolor[HTML]{FFF2CC}58.42 / 67.30} &
  \cellcolor[HTML]{FFF2CC}52.82 / 63.42 \\
\rowcolor[HTML]{EFEFEF} 
Ours \emph{(PV+YOLO+GDino)} &
  \multicolumn{1}{r|}{\cellcolor[HTML]{FFDB71}67.28 / 73.97} &
  \multicolumn{1}{r|}{\cellcolor[HTML]{FFDB71}60.36 / 68.01} &
  \cellcolor[HTML]{FFDB71}54.31 / 63.58 \\ \hline
\end{tabular}%
}
\caption{Perfomance on the Average Precision (AP) metric for our method on all difficulty sets on KITTI-val for the \emph{Pedestrian} class. Our late fusion method consistently outperforms the original specialized methods in 3DAP across all the difficulty levels. The color coding in yellow emphasizes the degree of improvement on 3DAP over the baselines.}
\label{table:main_ped}
\end{table}

\subsection{Average Precision}
Table \ref{table:main} presents the Average Precision (AP) for 3D object detection across the \textit{Easy}, \textit{Moderate}, and \textit{Hard} KITTI sets on the \emph{Car} class, comparing individual LiDAR models, camera models, and their fusion combinations. 
Our late fusion method shows consistently improved performance compared to any single-modality models, outperforming PV-RCNN and TED-S in 3D AP. The median improvement is +0.52 across all 3D-AP tests and difficulty levels, with an interquartile range of +0.43 to +0.86. While this improvement is modest, the performance on this dataset is already near saturation, leaving very little margin for error reduction. Hence, modest gains like this are typical of current state-of-the-art methods on KITTI for the \textit{Car} class. To verify the generalizability of our method, we evaluate it on the \emph{Pedestrian} class using models for which labels were available (Table \ref{table:main_ped}). Given a larger scope of error reduction for this class, we see that our fusion method demonstrates a median improvement of +2.31 with an interquartile range of +1.40 to +2.93 across all difficult levels.

Late fusion with general and open-vocabulary camera models (YOLO-NAS and GroundingDINO) perform as well as the tested special-purpose camera model (MonoDETR).  This is surprising since the baseline 2D-AP of these detectors is much lower than the specially trained model. This provides evidence that general-purpose camera detectors can be used to enhance special-purpose LiDAR detectors. 

Although our primary goal is to show that our late fusion method improves scores relative to the baseline methods, the table also provides numbers for comparison fusion methods. Our approach provides higher scores using a simpler method. 


\section{Limitations}
Our late fusion method is targeted at reducing false positives, but does not seek to increase true positives. We thus require that the LiDAR (or primary) detector successfully finds most true positives. Similarly, since our approach relies on validating LiDAR detections with camera inputs, it may not perform well with camera models that produce few detections and is, therefore, more applicable to methods that can be tuned for a greater number.

\section{Conclusion}
In this paper, we propose a model-adaptable late fusion method that combines LiDAR and camera detections to enhance object detection performance for autonomous vehicles. Our approach focuses on reducing false positives by cross-validating LiDAR detections with camera inputs. Experiments on the KITTI dataset demonstrate that our method is exceptional at its primary goal of reducing false positives, and provides modest but consistent improvements in average precision. The method performs competitively with state-of-the-art comparison methods. 

Importantly, the method works equally well with relatively low precision general purpose camera detectors which were not specifically trained on this dataset or for these specific detection classes. This is a surprising result, which we hope sparks future research on using general-purpose 2D detectors to improve the special purpose 3D vehicle detectors which currently dominate the state of the art.

\section*{Acknowledgments}
\noindent Partial support to some authors for this work was provided by WISEautomotive through the ATC+ Program award from MOTIE Korea 20014264.

\bibliographystyle{plain} 
\bibliography{IEEEexample}

\end{document}